%% file: main.tex
\def\year{2021}\relax
\documentclass[letterpaper]{article} 
\usepackage{aaai21_cameraready}  
\usepackage{times}  
\usepackage{helvet} 
\usepackage{courier}  
\usepackage[hyphens]{url}  
\usepackage{graphicx} 
\usepackage{natbib}  
\usepackage{caption} 
\usepackage[switch]{lineno}
\usepackage{amsmath}
\usepackage{amssymb}
\usepackage{multirow}
\title{Exploiting Relationship for Complex-scene Image Generation}
\author{
Tianyu Hua\textsuperscript{\rm 1}, 
    Hongdong Zheng\textsuperscript{\rm 1},
    Yalong Bai\textsuperscript{\rm 1},
    Wei Zhang\textsuperscript{\rm 1}\footnote{Corresponding author},
    Xiao-Ping Zhang\textsuperscript{\rm 2},
    Tao Mei\textsuperscript{\rm 1} \\
}
\affiliations{
    \textsuperscript{\rm 1}JD AI Research\\
    \textsuperscript{\rm 2}Ryerson University\\
    patrickhua.ty@gmail.com,
    \{hongdongzheng, ylbai\}@outlook.com,
    wzhang.cu@gmail.com\\
    xzhang@ee.ryerson.ca,
    tmei@live.com

}

\begin{document}

\maketitle

\input content/0abstract
\input content/1introduction

\input content/2relatedwork
\input content/3method

\input content/4experiment

\input content/5conclusion
\section*{Acknowledgments}
This work was supported by the National Key R\&D Program of China under Grant No. 2020AAA0108600, and 2020AAA0103800.
\input content/impact
\bibliography{reference}
\end{document}

%% file: content/0abstract.tex
\begin{abstract}

The significant progress on Generative Adversarial Networks (GANs) has facilitated realistic \textit{single-object} image generation based on language input. However, \textit{complex-scene} generation (with various interactions among multiple objects) still suffers from messy layouts and object distortions, due to diverse configurations in layouts and appearances. Prior methods are mostly object-driven and ignore their inter-relations that play a significant role in complex-scene images. This work explores relationship-aware complex-scene image generation, where multiple objects are inter-related as a scene graph. With the help of relationships, we propose three major updates in the generation framework. First, reasonable spatial layouts are inferred by jointly considering the semantics and relationships among objects. Compared to standard location regression, we show relative scales and distances serve a more reliable target. Second, since the relations between objects significantly influence an object's appearance, we design a relation-guided generator to generate objects reflecting their relationships. Third, a novel scene graph discriminator is proposed to guarantee the consistency between the generated image and the input scene graph. Our method tends to synthesize plausible layouts and objects, respecting the interplay of multiple objects in an image. Experimental results on Visual Genome and HICO-DET datasets show that our proposed method significantly outperforms prior arts in terms of IS and FID metrics. Based on our user study and visual inspection, our method is more effective in generating logical layout and appearance for complex-scenes.

\end{abstract}

%% file: content/1introduction.tex
\section{Introduction}

In the past few years, text-to-image generation has drawn extensive research attention for its potential applications in art generation, computer-aided design, image manipulation, etc. However, such success is only restricted to simple image generation, which only contains a single object in a small domain, such as flowers, birds, and faces \cite{reed2016generative,bao2017cvae}. Complex-scene generation, on the other hand, targets for synthesizing realistic scene images out of complex sentences depicting multiple objects as well as their interactions. Nevertheless, generating complex-scenes on demand is still far from mature based on recent studies~\cite{johnson2018image, xu2018attngan, li2019object,hinz2019generating}.

\begin{figure}[!t]
\centering
\includegraphics[width=0.9\linewidth,page=1]{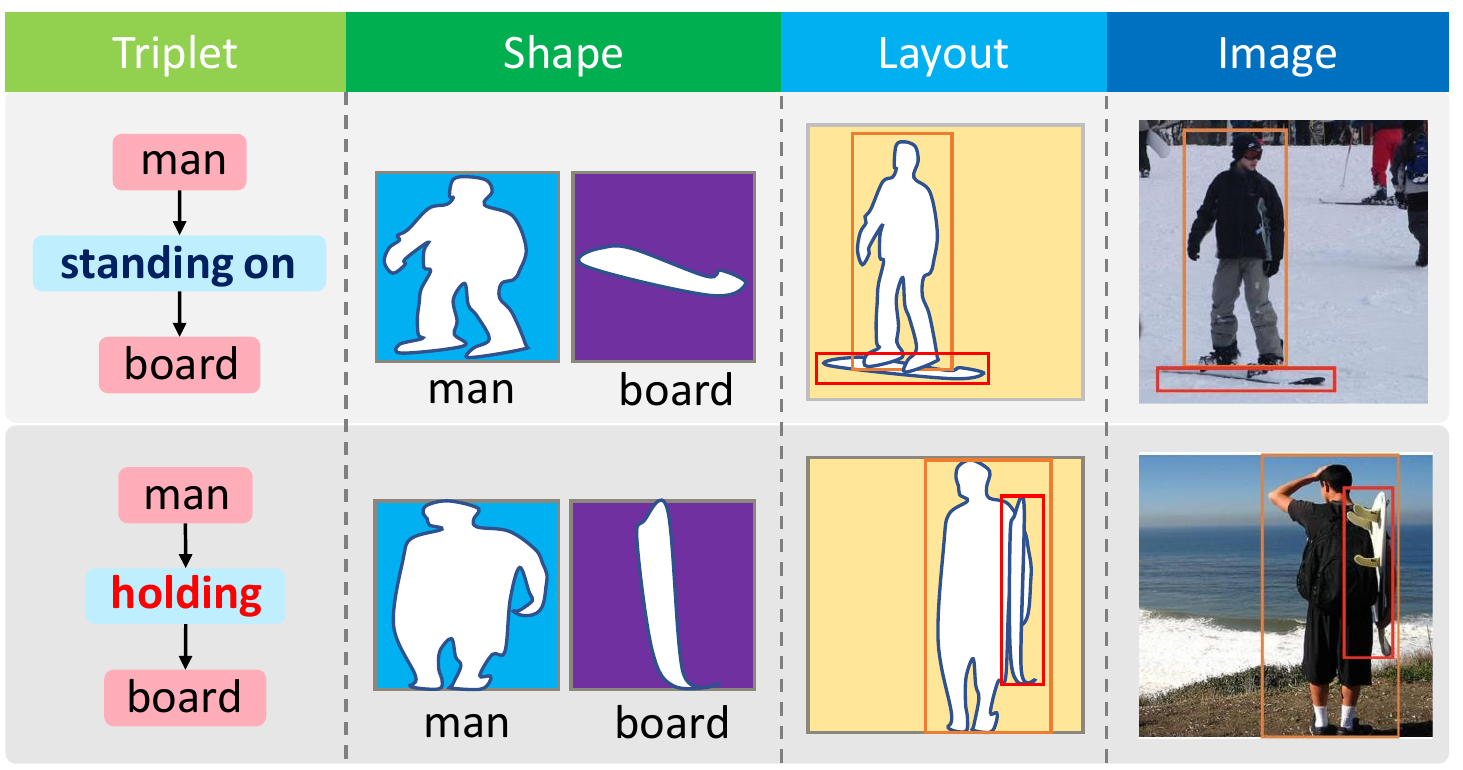}
\caption{Relationship matters for complex-scene image generation. The same object pair (e.g., \textit{man and board}) could show different object shapes, scene layouts and appearances under different relationships.} 
\label{fig:intro_cases}
\end{figure}

Scene graph, a structured language representation, captures objects and their relationships described in the sentence~\cite{xu2017scene}. Such representation is proven effective for image-text cross-modal tasks, such as structural image retrieval~\cite{johnson2015image,schuster2015generating,johnson2018image}, image captioning~\cite{yang2019auto,li2019know,li2018jointly} and visual question answering~\cite{teney2017graph,norcliffe2018learning}. 
In this work, we focus on complex-scene image generation from scene graphs. Although extensive works have been done in scene graph generation from image~\cite{xu2017scene,zellers2018neural,msdn,vtrans} (i.e. image$\rightarrow$scene graph), reversely generating a complex-scene image from a scene graph remains challenging, due to the polymorphism nature of one-to-many mapping from a given scene graph to multiple reasonable images with different scene layouts.

A general pipeline for scene graph based image generation usually consists of two stages~\cite{johnson2018image}. The first one learns to synthesize a rough layout prediction from the scene-graph input. Usually, the object features are encoded with a graph module~\cite{johnson2018image, ashual2019specifying}, followed by a direct regression of bounding-box locations. At the second stage, a position-aware feature tensor, that combines object features and layout generated in the first stage, is fed into an image decoder for generating the final image. For enhancing the object appearances in generated images, Ashual~\textit{et al.} separates appearance embedding from layout embedding. 

However, previous works on complex-scene generation heavily suffer from two fundamental problems: messy layout, and object distortion.
1) \textit{Messy layouts}. Image generation models are expected to figure out the reasonable layout from scene-graph inputs. However, there exist an infinite number of reasonable layouts for a given scene-graph. Directly fitting a specific layout introduces huge confusion, and limits the generalization ability. As a result, existing methods are still struggling with messy layouts in practice. 
2) \textit{Distortion in object appearance}. Due to the high diversity in object categories, layouts, and relationship dependencies, objects are often distorted during generation. For each object, the texture and local appearances should be inferred, respecting both its category and allocated spatial arrangement. Moreover, complex and various relations among different objects in the scene-graph can further increase the diversity of shape appearances. As shown in Fig.~\ref{fig:intro_cases}, even with the same object pairs, equipping different relationships could lead to totally different scene layouts and appearances. 

Some works~\cite{ashual2019specifying} simplify the task by only taking a few simple spatial relationships among objects (such as ``left'', ``right'' or ``above'') but ignoring other complex relationships (such as verbs). Meanwhile, to reduce the complexity, some works only consider one specific stage of this task, such as layout generation from scene-graph~\cite{jyothi2019layoutvae}, image generation from layout~\cite{zhao2018image, sun2019image}. All these works did not take account of the semantics and complex relationships among objects, which limits their application prospects.

\begin{figure}[!t]
\centering
\includegraphics[width=0.99\linewidth]{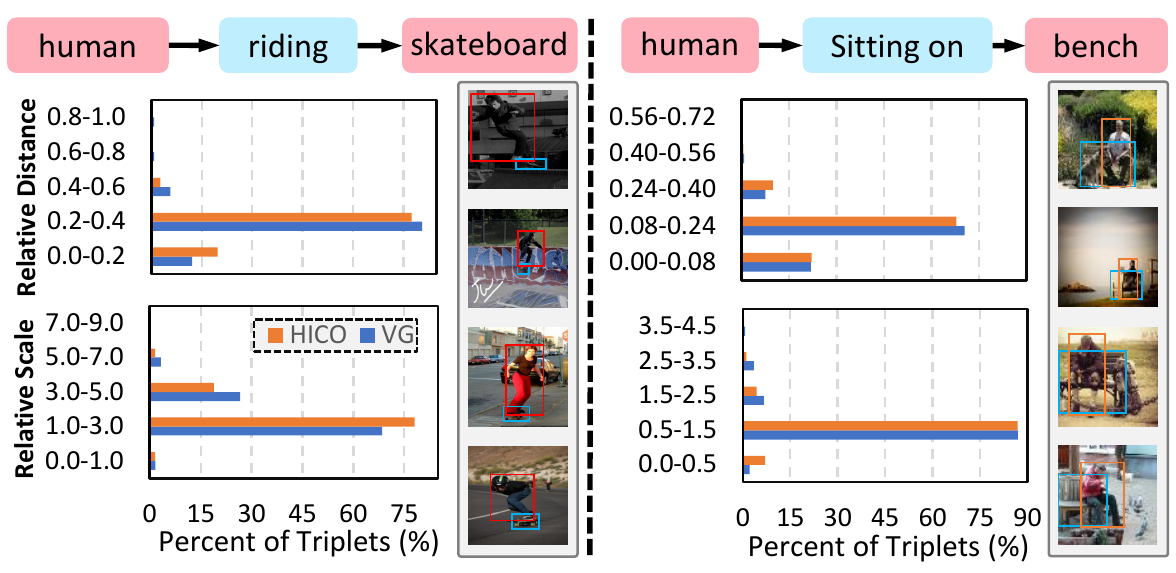}
\caption{Distributions of relative scale and distance for ``man riding  skateboard'' and ``man sitting~on bench''.}
\label{fig:scale_dist_example}
\end{figure}

In this work, we explore relationships to mitigate the above issues. We consider both simple spatial relationships and complex semantic relationships into consideration. We observed that, in different realistic images, relative scale and distance ratios between two interrelated objects from the same ``subject-relation-object'' triplet usually conform to a norm distribution with low variance, as in Fig.~\ref{fig:scale_dist_example}. Even though the ``human'' have various poses, and the skateboard can be oriented to different directions, the scale ratio between the two bounding boxes is naturally clustered with very low variance. Thus, we first introduce relative scale ratios and distance for measuring the rationality of layouts generated from the scene graph. It means that all \textit{various reasonable layouts} relevance to one specific scene graph can be measured under a common standard and result in very similar results. After that, we proposed a \textit{Pair-wise Spatial Constraint Module} for assisting layout generation. Our Spatial Constraint Module is influenced by object pairs and their corresponding relation jointly. Meanwhile, the objective of this novel module is to correct the layout by fitting the relative scale ratio and relative distance ratio between interrelated object pair beside the absolute position coordinates of each object. In this way, the spatial commonsense of complex scene with multiple objects can be modeled.

Moreover, for enhancing the influence of relation for object appearance generation, we proposed a \textit{Relation-guided Appearance Generator} and a novel \textit{Scene Graph Discriminator} for guiding image generation. Unlike the traditional discriminator that only judges whether the image is fake or not, our proposed new discriminator has two main functions. One is to determine whether the objects in the generated image are relevant to the objects described in the text scene graph or not, and the other is to discriminate the relation predictions among objects from the generated image are correspondence with the relationship described in the input scene graph. By feeding back these strong discriminant signals, our Scene Graph Discriminator guarantees the generated object patches align with not only single object fine-grained information but also the relation discrepancy among objects. 

The main contributions can be summarized as follows:
\begin{itemize}
\item A novel pair-wise spatial constraint module with supervisions of relative scale and distance between objects for learning relationship-aware spatial perceptions. 
\item A relation-guided appearance generator module followed by a scene graph discriminator for generating reasonable object patches respecting object fine-grained information and relation discrepancy.
\item A general framework for synthesizing scene layout and images from scene graphs. The experimental results on Visual Genome~\cite{krishna2017visual} and human-objects interactions dataset HICO-DET~\cite{chao2018learning} demonstrate the complex-scene images generated by our proposed method follow the common sense.
\end{itemize}

%% file: content/2relatedwork.tex
\section{Related Work}


\textbf{Image Synthesis from Sentence} is a conditional image generation task whose conditional signal is natural language. Textual descriptions are traditionally fed directly to a recurrent model for semantic information extractions. After that, a generative model will produce the results conditioned on this vectorized sentence representation. Most of these tasks specialize in single object image generation~\cite{reed2016generative,zhang2017stackgan,xu2018attngan}, whose layout is simple and the object usually centered with a large area in the image. However, generating realistic multi-object images conditioned on text descriptions is still a difficult task, since it addresses very complex sense layout generation and various object appearances mapping, and both of scene layout and object appearances are heavily influenced by the spatial and semantic relationships cross objects. Furthermore, encoding all information, including multiple object categories and the interactions among them into one vector, usually leads to critical details lost. Meanwhile, directly decoding images from such an encoded vector hurts the interpretability of the model.

\textbf{Scene Graph}~\cite{xu2017scene} is a directed graph that represents the structured relationships among objects in an explicit manner. Scene graphs have been widely used in many tasks such as image retrieval~\cite{johnson2015image}, image captioning~\cite{anderson2016spice}, which serves as a medium that bridges the gap between language and vision. 


\textbf{Image Synthesis from Scene Graph}~\cite{johnson2018image} is a derivative of sentence based multiple-object image generation. Since the conditional signals are scene graphs, graphic models are usually applied for extracting essential information from the textual scene graph. After that, these extracted features are directly used for regressing the scene layouts and then treated as input to an image decoder for generating the final image~\cite{ashual2019specifying}. Such a framework is applicable to generation image contains multiple objects with simple spatial interactions. However, it is still suffering from modeling the reasonable scene layouts and appearances following commonsense due to the implication of semantic relationships in the scene graph.


In this paper, we focus on image generation from the textual scene graph. Different from previous methods, we highlight the impact of relationships among objects for dealing with the messy layout and various object appearance.


%% file: content/3method.tex
\section{Method}
A scene graph is denoted as $\mathcal{G}=\{ \mathcal{C}, \mathcal{R}, \mathcal{E} \}$, where $\mathcal{C}=\{c_1, c_2, ..., c_n\}$ indicate the nodes in the graph, each $c_i\in \mathcal{C}$ denotes the category embedding of an object or instance. Note that we use words like "object" or "instance" in reference to a broad range of categories from "human", "tree" to "sky", "water" etc. The edges of the graph are extracted as a relationship embedding set $\mathcal{R}$. Two related objects $c_j$ and $c_k$ associate with one relationship $r_{jk}\in \mathcal{R}$, which leads to a triplet $\left \langle c_j, r_{jk}, c_k \right \rangle$ in the directed edge set $\mathcal{E}$.

\begin{figure*}[!ht]
\centering
\includegraphics[width=0.95\linewidth,page=1]{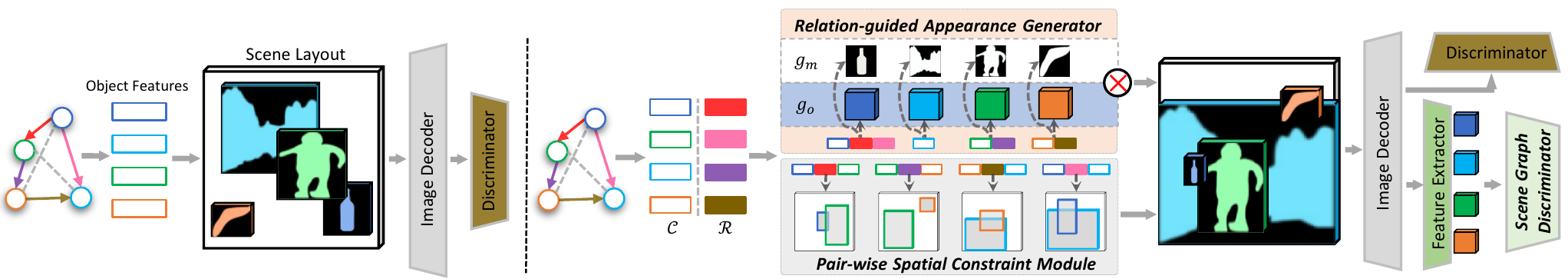}
\caption{Illustrations of standard (left) and our (right) framework for scene graph to image generation. Left: Directly generating layout and image based on object features extracted from scene graph. Right: Our proposed framework with object pair-wise spatial constraints and appearance supervision respecting relationships among objects.}
\label{fig:intro_pipeline}
\end{figure*}

Given a scene graph $\mathcal{G}$ and its corresponding image $I$, scene graph-based image generation model aims to generate an image $\widehat{I}$ according to $\mathcal{G}$ by minimizing $D(I, \widehat{I})$, where $D(I, \widehat{I})$ measures differences between $I$ and $\widehat{I}$. A standard scene graph to image generation task can be formulated as two separate tasks: a scene graph to layout generation task which extracts object features with spatial constraints from scene graphs, and an image generation task, which generates images conditioned on the predicted object features and learned layout constraints, as shown in Fig.~\ref{fig:intro_pipeline} (left).

In this paper, we extend the traditional framework by emphasizing the influence of relationship $\mathcal{R}$ for both object layouts and object appearances generation. As shown in Fig.~\ref{fig:intro_pipeline} (right), three novel modules are proposed:
\begin{itemize}
    \item \textbf{Pair-wise Spatial Constraint Module}: a module for constraining layout generation according to the semantic information extracted from $\mathcal{E}$. 
    \item \textbf{Relation-guided Appearance Generator}: for each object $c_i$, we introduce the semantic information of its connected relationships $\{r_{j}|\left \langle c_i, r_{j}, * \right \rangle\in\mathcal{E}\}$ to influence the shape and appearance of the generated image of $c_i$.
    \item \textbf{Scene Graph Discriminator} ($D_{sg}$): a novel discriminator for strengthening the generated image $\widehat{I}$ to be relevant to the appearances of object $\mathcal{C}$, and the relationships $\mathcal{R}$ in the edge set $\mathcal{E}$.
\end{itemize}

\subsection{Layout Generator}
Layout generator aims to predict bounding boxes $b_i=(x_i,y_i,w_i,h_i)$ for each object $o_i$ in given scene graph $\mathcal{G}$, where $x_i,y_i,w_i,h_i$ specifies normalized coordinates of the center, width and height in ground-truth image $I$.

In previous work, the object representations are usually extracted from scene graph inputs, and then be passed to a box regression network to get the bounding box predictions $\widehat{b}_{i} = (\widehat{x}_i, \widehat{y}_i, \widehat{w}_i, \widehat{h}_i)$. The box regression network is trained by maximizing the objective:
\begin{equation}
    \mathcal{L}_{box} = -\sum_{i=1}^n \parallel b_i - \widehat{b}_i \parallel_2,
\end{equation}
which penalize the $L_2$ difference between ground-truth and predicted boxes. $n$ indicates the amount of objects.

Since there are various reasonable layouts existing, as previously stated, a scene graph to layout task requires addressing challenging one-to-many mapping. Directly regressing layout to offsets of one specific bounding box would hurt the generalization ability of the layout generator, and make the layout generator to be difficult to convergence. In order to generate reasonable layouts, we relax the constraint of bounding box offsets regression and proposed a novel spatial constraint module for ensuring the rationality of layout.

Our \textbf{Pair-wise Spatial Constraint Module} introduces two novel metrics for measuring the realistic of layouts.

\begin{figure}[!t]
\centering
\includegraphics[width=0.98\linewidth]{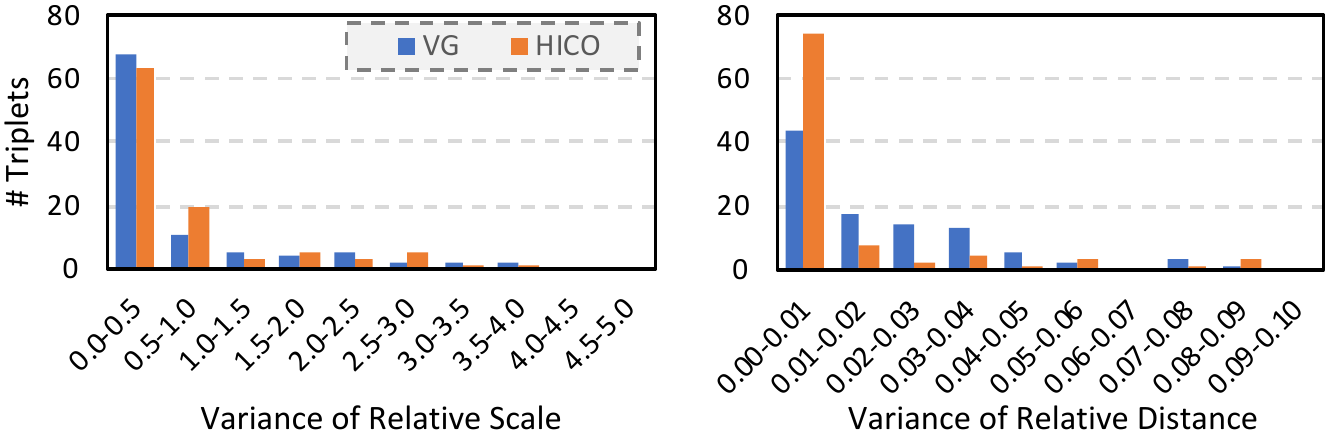}
\caption{Distributions of relative scale and distance variance among top-100 triplets in VG and HICO-DET datasets. Low diversity of relative scale and distance is observed, following the property of commonsense knowledge.}
\label{fig:scale_dist_variance}
\end{figure}

\noindent\textit{1. Relative Scale Invariance}. The scale of an object is represented by the diagonal length of its bounding box. For any given $\left \langle c_j, r_{jk}, c_k\right \rangle$ triplet, the ratio between the scale of the subject and the scale of the object in different images are often roughly the same, as shown in Fig~\ref{fig:scale_dist_variance} (Left). We formulate the relative scale between the layout $b_j$ and $b_k$ as
    \begin{equation}
        s_{jk} = {\sqrt{w_j^2+h_j^2}}\Big/{\sqrt{w_k^2+h_k^2}}.
    \end{equation}
\noindent\textit{2. Relative Distance Invariance}. Similar to relative scale, relative distance target at calculating the distance between two objects in triplet normalized by the scales of two objects. The relative distance of related object pair $c_j$ and $c_k$ in realistic images is also naturally clustered to one specific value, and the distributions of relative distance for different triplets are usually with low variance, as shown in Fig~\ref{fig:scale_dist_variance} (Right). Normally, horizontal flips of images rarely alter spatial relationship distributions, we relax this constraint by using the absolute value of the horizontal coordinate difference. Most importantly, we normalize distance by the summed scales of object pairs so that the zooming effect of object depth can be canceled out. Therefore, the relative distance between the layout $b_j$ and $b_k$ can be formulated as
    \begin{equation}
    \Vec{d_{jk}} = {[|x_j-x_k|, y_j-y_k]^T}\Big/{\left(\sqrt{w_j^2+h_j^2}+\sqrt{w_k^2+h_k^2}\right)}.
    \end{equation}


We have keenly observed that relationship in a semantic form comes with it an inherent spatial constraint that has not been fully explored by others. For example, the relationship ``holding'' implies that the object should be within arm's reach of the subject instead of miles away. The relationship ``walking'' indicates the relative vertical arrangement between subject and object heavily, whether it's ``man walking-on street'' or ``dog walking-on floor''. It means the relative scale and relative distance between two objects heavily depend on the relationship or interaction between these two objects. Therefore, we devise a training scheme that explicitly leverages this constraint. 

In this work, the scene graph $\mathcal{G}$ is first converted to object feature vectors $\mathcal{C}$ and relation embeddings $\mathcal{R}$, and then fed into a Graph Convolutional Network (GCN). The GCN outputs updated object level feature vector $o_i = T(c_i,\mathcal{C}_i, \mathcal{R}_i)$ aggregated with relation information, where $T$ is the graph convolutional operation, $\mathcal{C}_i$ is the set of object embeddings relevant to $c_i$, $\mathcal{R}_i$ is the set of embeddings for relations among $c_i$ and $\mathcal{C}_i$. It means the output vector $o_i$ for an object $c_i$ should depend on representations of relationships and all objects connected via graph edged jointly. After that, we apply the updated object representations for generating the layout for object $c_i$ by $\widehat{b}_i = B(o_i)$, where $B$ is an bounding box offset regression network. We construct a scale-distance objective for our proposed spatial constraint module to assist the training progress of $B$:
\begin{equation}
    \mathcal{L}_{scm}\!=\!-\sum_{\substack{0<j,k<n \\ \left \langle c_j, r_{jk}, c_k \right \rangle\in\mathcal{E}}}\parallel s_{jk} - \widehat{s}_{jk} \parallel_2\!+\!\parallel \Vec{d}_{jk} - \Vec{\widehat{d}_{jk}}\parallel_2,
\end{equation}
where $\widehat{s}_{jk}$ and $\Vec{\widehat{d}_{jk}}$ is the relative scale and relative distance between generated layouts for related object pair $c_j$ and $c_k$ respectively. $\mathcal{L}_{scm}$ is only computed on the connected object pairs in scene graph, since the relative scale and distance of two objects depend on the relationship between them, as we shown in Fig.~\ref{fig:intro_cases}. 

With the supervision of relative scale and distance, the box regression network learns to arrange object boxes properly for reasonable layout generation.

\subsection{Image Generator}
Starting from the original object representations $\mathcal{C}\in\mathbb{R}^{n\times d_1}$ and initial relation embeddings $\mathcal{R}\in\mathbb{R}^{m\times d_2}$, we can compute a combined ``object-relation'' vector $v_i$ for each object $c_i$ in scene graph:
\newcommand{\concat}{\ensuremath{+\!\!\!\!+\,}}
\begin{equation}
    v_i = \Big(c_i\concat\frac{1}{|\mathcal{E}_i|}\sum_{\left \langle c_i, r_j,*\right\rangle \in \mathcal{E}_i} r_j\Big) + z_i,
\label{rel_guided}
\end{equation}
where $\concat$ indicates a vector concatenation operation, $\mathcal{E}_i\in\mathcal{E}$ is the collection of all triplet relevant to object $c_i$, $z_i$ is a $d_1+d_2$ dimensional noise vector randomly sampled from a Gaussian distribution, which aims to generate non-deterministic object features. The object and averaged relation embeddings are eventually be concatenated as inputs of our \textbf{Relation-guided Appearance Generator}, which consists of an object mask predictor $g_m$, an object appearnce feature predictor $g_o$ and a full image generator.

The combined vector $\{v_i\}_{i=1}^n$ will be sent simultaneously to $g_m$ and $g_o$, both of which are four-layer conv nets normalized with spectral normalization techniques~\cite{miyato2018spectral}. Through an STN block~\cite{spatial2015Jaderberg}, the two outputs for different objects will first be filled into their respective bounding box layouts. Then we obtain a set of object shape tensor and appearance tensor. By multiplying these two tensor, we can generate the final \textit{relation-guided} appearance feature tensor for all objects in scene graph as 
\begin{equation}
    a(\mathcal{G}) = \{S(\widehat{b}_i, g_m(v_i))\circ S(\widehat{b}_i, g_o(v_i))\}_{i=1}^{n},
\end{equation}
where $S$ is the STN block.

After that, our full image generator generate the image conditioned on all object appearance feature tensors $a(\mathcal{G})$ and an additional noise vector $z_{I}$. In detail, our image generator utilizes the ResNet architecture~\cite{he2016deep} consists of six ResBlocks as backbone. Consider generating a 256$\times$256 image for scene graph, a randomly generated global latent vector $z_{I}$ is a vector sampled from normal distribution. The vector is then mapped and reshaped to a 1024$\times$4$\times$4 (channels, width, and height) tensor through a fully-connected layer. Then, the tensor will be sent to the first ResBlock. Each of the six ResBlocks will upsample it's inputs bilinearly with a ratio of two. In the meantime, the channel number drops by a factor of 2 except for the third block. Block by block, we fuse object appearance tensor $\{f_{i} = G_{obj}(v_i)\}_{i=1}^n$ with the outputs of each ResBlock (global appearance tensor) using the ISLA-Norm method proposed by \cite{sun2019image}. The final generated image $\widehat{I} = \widehat{I}_t$ comes from the outputs of the last ResBlock, 
\begin{equation}
\begin{split}
    \widehat{I}_t &= R_t(\widehat{I}_{t-1},a(\mathcal{G})) \\
    \widehat{I}_1 &= R_1(z_I,a(\mathcal{G}))
\end{split}    
\end{equation}
where $R$ indicate a ResBlock equipped with ISLA-Norm module, $t$ is the amount of ResBlocks in our image generator, $\widehat{I}_i$ is output of the $i$-th ResBlock. 

Our image generator takes the object appearance features with relational information and global random noise as condition, adapts scene composition, and finally generates the realistic image $\widehat{I}$ for scene graph $S$.

\subsection{Image Discriminator}

Similar to the image generator, we adapt ResNet with downsampling blocks for image discriminator. The ResNet backbone consists of a different number of downsampling ResBlocks with respect to the input image sizes. 
The image downsampled by ResBlocks goes through a linear layer, and the outputs of the linear layer are further summed channel-wise to form a scalar output as the global discriminator score $D_{img}$ to measure whether the input image is real or not, which is similar to traditional GAN based methods.

Since different relationships result in diverse appearance in the same object, we argue that the learned object feature representation reflects not just class-related object styles but also the relationship-aware appearances. Thus, we proposed a novel \textbf{Scene Graph Discriminator} $D_{sg}$ to measure whether the scene graph extracted from the generated image is associated with the given textual scene graph or not. In detail, we first extract object-level feature patches $\{p_i\}_{i=1}^n$ rerouted from the second layer of ResNet backbone, then resize these feature patches to the same size by an RoI align layer~\cite{he2017mask}. Then we introduce an object classifier $F_{obj}$, which attempts to classify the feature patches into categories. By pairing object feature tensors according to the edges of the scene graph, we send the paired object feature $p_j$ and $p_k$ to the relationship classifier $F_{rel}$, which aims to predict the type of relationship of given object feature pair. Our proposed $D_{sg}$ aims to encourage the image generator to be aware of the object categories and relationships exists in the scene graph:
\begin{equation}
    D_{sg}(I)\!=\!\frac{1}{n}\sum_{i = 0}^{n}F_{obj}(c_i|p_i)+\frac{1}{|\mathcal{E}|}\!\sum_{\substack{0<j,k<n \\ \left \langle c_j, r_{jk}, c_k \right \rangle\in\mathcal{E}}}\!F_{rel}(r_{jk}|p_j,p_k).
\end{equation}
Moreover, we introduce an object discriminator $D_{obj}$ to measure whether each object in image appears realistic based on $\{p_i\}_{i=1}^n$.

The overall objective function for training layout generator, image generator and discriminators is defined as:
\begin{equation}
    \mathcal{L} = \lambda_{1}\mathcal{L}_{box} + \lambda_{2}\mathcal{L}_{scm} +  \lambda_{3}\mathcal{L}_{obj} + \lambda_{4}\mathcal{L}_{sg} + \lambda_{5}\mathcal{L}_{img},
\end{equation}
where $\mathcal{L}_{img}$ is image adversarial loss from $D_{img}$,  $\mathcal{L}_{obj}$ is object adversarial loss from $D_{obj}$, $\mathcal{L}_{sg}$ is scene graph relevant loss from $D_{sg}$. In our experiments, we set the loss weight parameters $\lambda_{1},\lambda_{2},\lambda_{3},\lambda_{4}=1$, $\lambda_{5} = 0.1$.

%% file: content/4experiment.tex
\section{Experimental Results}


We evaluate our proposed method for generating images at three different resolutions 64$\times$64, 128$\times$128, and 256$\times$256 in below two datasets:

\noindent\textbf{Visual Genome} \cite{krishna2017visual} was constructed with cognitive tasks that provide crowd-sourced dense annotations of both scene graphs and images. Following the settings of \cite{johnson2018image}, we experiment on Visual Genome version 1.4. 
We keep 178 objects and 45 relations in the dataset by removing images with object and relationship categories less than 2000 and 500, respectively. 

\noindent\textbf{HICO-DET}~\cite{chao2018learning} was built for modeling humans-object interactions. Compared with Visual Genome, the scene graphs provided in the dataset are human-centered. 
We keep images that have object categories higher than 1000 and discard images with interaction types that repeat less than 250 times, leaving 19 objects and 22 relationship types in total. Images with Object size below 32$\times$32 and images with objects less than 2 or more than 10 are ignored. Finally, we get 15963 train images and 4034 test images.

The COCO dataset~\cite{caesar2018coco-stuff} is not used in this paper because the relationship types in COCO are too simple, consisting mainly of naive spatial arrangement relations. 
We trained models using Adam with an initial \textit{lr}=$10^{-4}$ and batch size of 32 for 200 epochs. 

Several previous works target at multi-object image generation. Most of these works are about image synthesis from the ground-truth layout or pixel-level instance segmentation annotation~\cite{sun2019image,hong2018inferring,li2019object}. The work of \citeauthor{ashual2019specifying} aims to generated images from an input scene graph. However, the scene graph used in their work is simplified by only equipping six spatial relationships (right-of, left-of, above, below, surrounding and inside). Moreover, location attributes are assisted by additional information for each node in scene graph. Moreover, location attributes are assisted by additional information for each node in scene graph. \citeauthor{luo2020end} only focus on spatial relationships instead of semantic relationships. Besides, objects used in their paper are mostly rigid bodies. Our paper learns from not just spatial relationships, but semantic relationships (e.g. looking at) as well. We use datasets that involve a large number of non-rigid objects that have various shapes and appearances and be sensitive to their relevant semantic relationships, which drastically increase the difficulty of our task. 
PasteGAN~\cite{yikang2019pastegan} applies both of the scene graph and ground-truth image crops as the inputs for complex-scence image generation. According to our best knowledge, sg2im~\cite{johnson2018image} is the only related work about complex-scene image generation images from scene graphs that contain semantic and complex relationships among objects. 

\noindent\textbf{Compared Methods} In this paper, we compare our proposed method with \textbf{sg2im} and \textbf{PasteGAN} for complex-scene image generation. Moreover, to demonstrate the effectiveness of our relation-guided appearance generator and scene graph discriminator, we also compare our method with \textbf{LostGAN}~\cite{sun2019image} which is designed for generating images by given ground-truth layout.


\begin{table}[!t]
\small
\centering
\setlength{\tabcolsep}{1.4mm}{
\begin{tabular}{l|c|cc|cc}
\hline
\multirow{2}{*}{Resolution}& \multirow{2}{*}{Method} &\multicolumn{2}{c|}{Visual Genome} & \multicolumn{2}{c}{HICO-DET}\\\cline{3-6}
  &  & IS & FID & IS & FID \\
\hline\hline

\multirow{7}{*}{64$\times$64} & $I$  & 13.9 $\pm$ 1 & 0.0 &  9.8 $\pm$ 0.5 & 0.0 \\
& sg2im$\dagger$ & 6.3 $\pm$ 0.2 &47.6 & 4.4 $\pm$ 0.1 & 99.9\\
& LostGAN$\dagger$ & 6.9$\pm$0.1 & 38.7 & 4.5$\pm$0.3 & 86.4 \\
& Ours$\dagger$ &  \textbf{7.5} $\pm$ \textbf{0.4}  & \textbf{29.0} &  \textbf{5.5} $\pm$ \textbf{0.1} & \textbf{41.7}\\ 
& sg2im & 5.5 $\pm$ 0.1 &47.5 & 4.4 $\pm$ 0.1 & 94.3\\
& PasteGAN & 6.9$\pm$0.2 & 58.5 & - & -\\
& Ours & \textbf{7.0} $\pm$ \textbf{0.2} & \textbf{37.7} & \textbf{5.3} $\pm$ \textbf{0.7} & \textbf{47.4}\\

\hline

\multirow{6}{*}{128$\times$128} & $I$  & 22.5 $\pm$ 1.9 & 0.0 &  13.7 $\pm$ 0.7 & 0.0 \\
& sg2im$\dagger$ & 6.3 $\pm$ 0.2 & 83.9 & 4.6 $\pm$ 0.1 & 83.7\\
& LostGAN$\dagger$ & 7.4$\pm$0.3 & 53.4 & 4.8$\pm$0.1 & 79.9 \\
& Ours$\dagger$ &  \textbf{9.4} $\pm$ \textbf{0.4}  & \textbf{41.0} &  \textbf{6.5} $\pm$ \textbf{0.1} & \textbf{60.6}\\ 
& sg2im & 6.2 $\pm$ 0.2 & 83.8 & 4.6 $\pm$ 0.1 & 123.0\\
& Ours & \textbf{9.2} $\pm$ \textbf{0.8} & \textbf{53.0} & \textbf{5.0} $\pm$ \textbf{0.3} & \textbf{61.4}\\
\hline

\multirow{3}{*}{256$\times$256} & $I$  &  30.1 $\pm$  2.3 & 0.0 & 16.3 $\pm$ 0.5 & 0.0 \\
& Ours$\dagger$ & 12.6 $\pm$ 0.5  & 68.3 &7.5 $\pm$ 0.1& 78.3\\ 
& Ours  & 10.8 $\pm$ 0.9 & 85.7 & 6.9 $\pm$ 0.3 & 80.5 \\
\hline
\end{tabular}}
\caption{The comparison of IS and FID among different methods. On each dataset, the test set samples are randomly split into 5 groups. The mean and standard deviation across splits are reported in the above table. $\dagger$ indicates that the images are generated based on the ground-truth layouts instead of the generated layouts. $I$ denotes the real image.}\label{tab:IS}
\end{table}

\begin{figure*}[!t]
\centering
\includegraphics[width=1\linewidth,page=1]{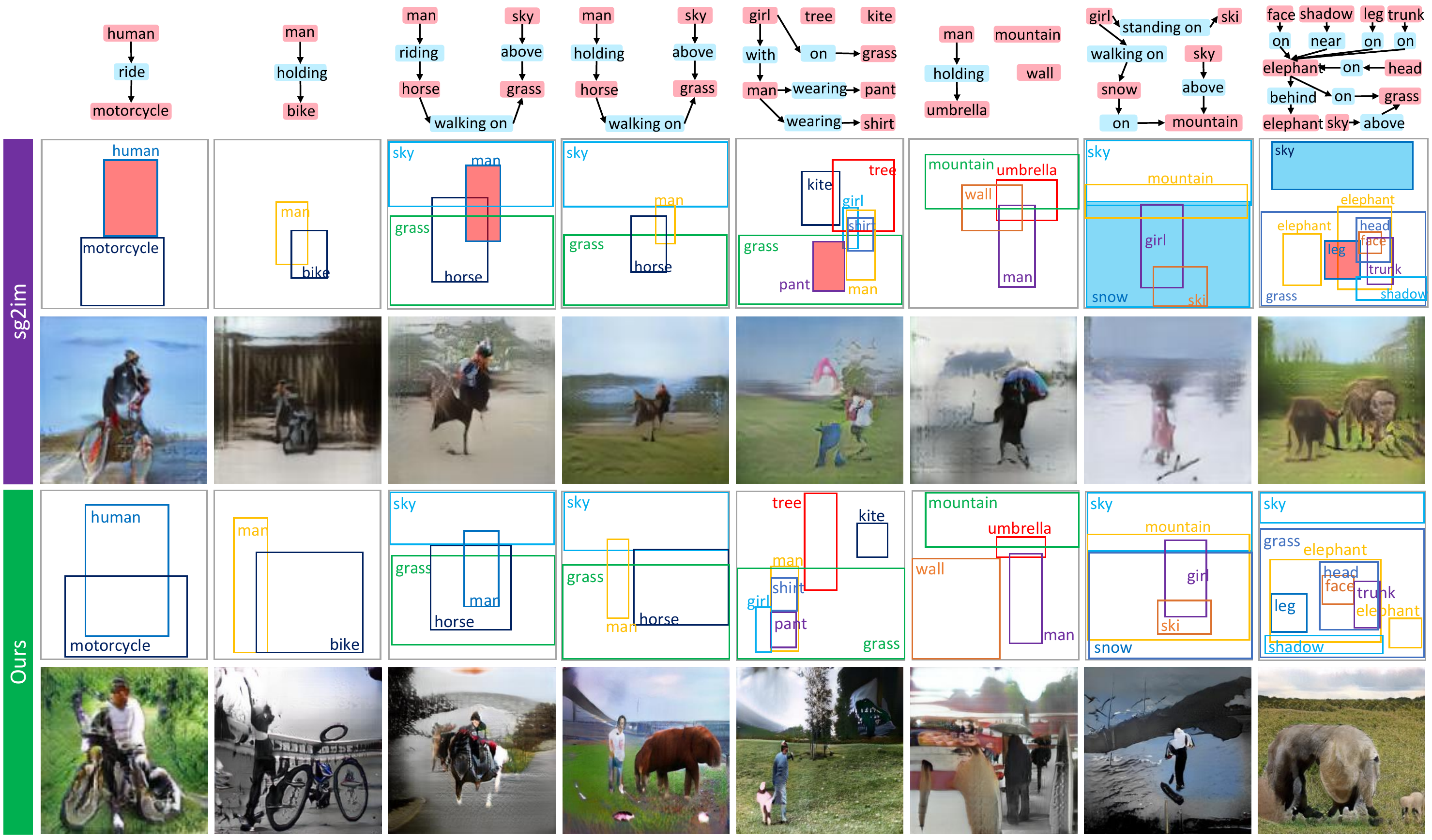}
\caption{Examples of layouts and images generated from scene graphs in Visual Genome and HICO-DET for our method and sg2im. In the layout examples, we use red color patches to denote bounding boxes that fail to reflect the distance between object pairs. The bounding boxes with blue background have an unnatural scale configuration. Best viewed in color version.}
\label{fig:cmp}
\end{figure*}

\subsection{Quantitative Results}

We adopt two metrics to evaluate the generated images. \\\textbf{Inception Score} (\textit{IS}) \cite{salimans2016improved} measures the diversity of generated images and their quality. A pre-trained InceptionV3 model is adapted to predict the class probabilities for given image. Larger inception scores are better.\\\textbf{Fréchet Inception Distance} (\textit{FID}) \cite{heusel2017gans}\footnote{https://github.com/mseitzer/pytorch-fid} measures the Fréchet distance between the multivariate Gaussian distribution of real images and generated ones. Lower \textit{FID} scores are better.\\
These two metrics are widely used evaluation metrics for generative models. \textit{IS} aims to evaluate the reality of a single object, while \textit{FID} is more suitable to reflect the quality of the generated image contains multiple objects.

Table~\ref{tab:IS} summarizes the performances on the two aforementioned datasets in terms of Inception Score and \textit{FID} score. Our model outperforms sg2im on both VG and HICO-DET datasets. 
Moreover, even without the external information like image crops, our method can still achieve better results reported in PasteGAN. 
In addition, we conduct experiments of GT Layout versions using ground-truth bounding boxes during both training and testing. This method gives an upper bound to the model's performance in the case of perfect layout construction. As shown in Table~\ref{tab:IS}, our method has more potential than sg2im and LostGAN.

We also conducted ablation studies in Table~\ref{tab:ablation}. The relative importance of the \textit{Pair-wise Spatial Constraint Module}, and \textit{Scene Graph Discriminator} are measured by removing $\mathcal{L}_{scm}$ and $\mathcal{L}_{sg}$ from the overall objective function respectively. The ablation studies of \textit{Relation-guided Appearance Generator} is measured by erasing relation embeddings during computing object shape and texture features. 
It can be found that the layout constraint module predicts reasonable spatial layout arrangements that improve the generated image qualities. The relation-guided generator introduces more reasonable appearance information. The scene graph discriminator can advance the correspondence between textual scene graph inputs and generated images. 

\begin{table}[!t]
\centering
\small
\setlength{\tabcolsep}{1.2mm}{
\begin{tabular}{c|c|c}
\hline
Method & IS & FID \\
\hline\hline
Ours & 9.2 $\pm$ 0.8 & 53.0\\
w/o Pair-wise Spatial Constrain Module ($\mathcal{L}_{scm}$) & 8.6 $\pm$ 1.2 & 59.8\\
w/o Relation-guided Appearance Generator & 8.7 $\pm$ 0.9 & 57.4\\
w/o Scene Graph Discriminator ($\mathcal{L}_{sg}$) & 7.4 $\pm$ 0.2 & 73.3\\
\hline
\end{tabular}}
\caption{Ablation studies conducted on our proposed method. The experimental results are reported on 128$\times$128 resolution image generation task in Visual Genome.}\label{tab:ablation}
\end{table}

\begin{figure}[!t]
\centering
\includegraphics[width=1\linewidth,page=1]{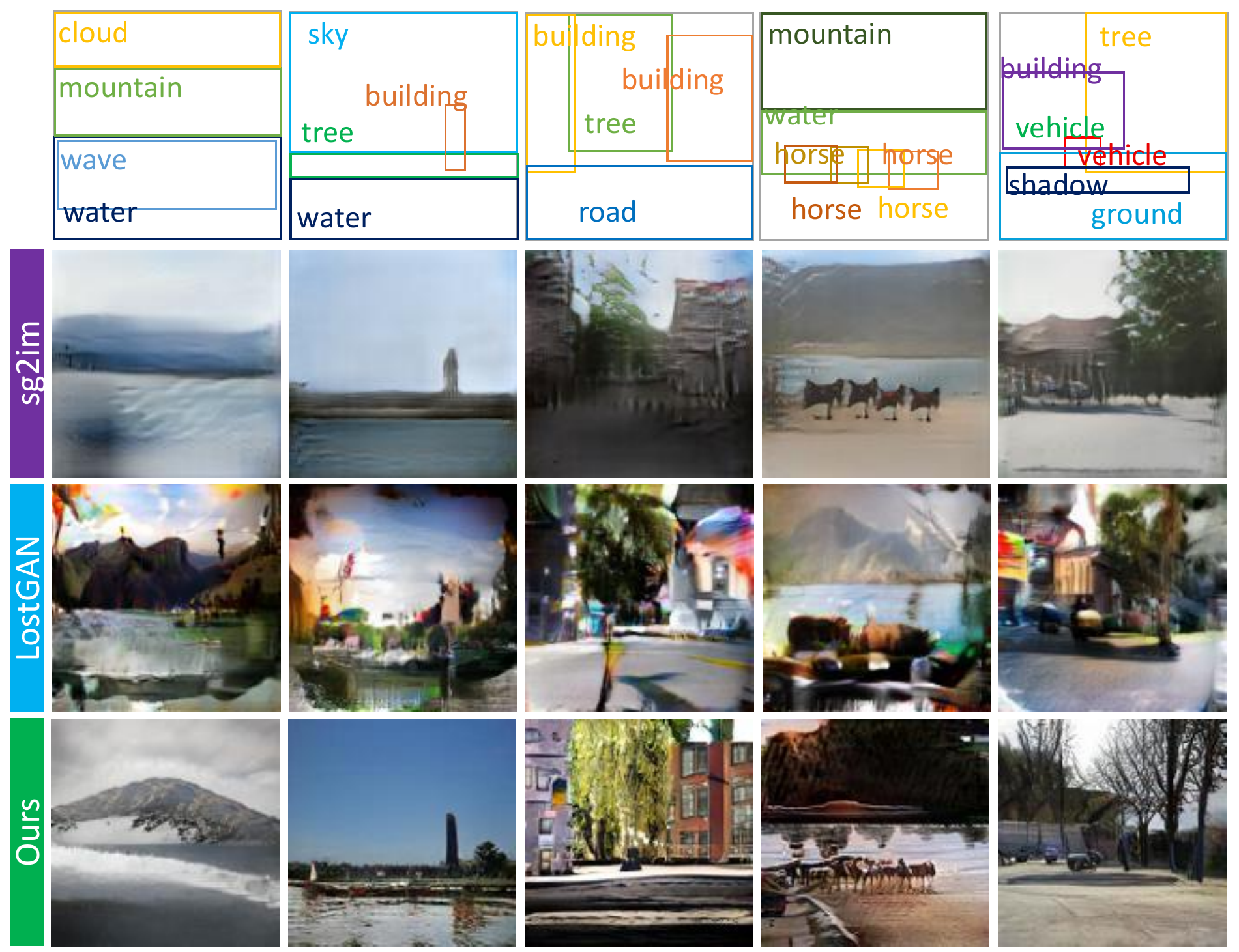}
\caption{Generated samples from ground truth layouts on Visual Genome by sg2im, LostGAN and our method.}
\label{fig:cmp_gt}
\end{figure}

\begin{table}[!t]
  \small
  \centering
  \setlength{\tabcolsep}{1.2mm}{
  \begin{tabular}{p{5.3cm}|l|l|l}
  \hline
  User study                                                              & sg2im & Same & Ours \\ \hline\hline
  Image is more realistic                                          &  9\% & 26\% & 65\%  \\ \hline
  Image has reasonable object arrangement                          & 12\% & 27\% & 61\%  \\ \hline
  Image reflects relationships in scene graph                  &  9\% & 19\% & 72\%  \\ \hline
  Layout is more reasonable                                        & 11\% & 30\% & 59\%  \\ \hline
  \end{tabular}}
  \caption{We performed a user study to compare the quality of generated layouts and images of our method against sg2im.}
  \label{tab:user_study}
  \end{table}

\subsection{Qualitative Results}
Fig.~\ref{fig:cmp} shows the capability of our method compared with that of sg2im on VG and HICO-DET. 
In the 1st column, sg2im predicts the human layout is above the motorcycle, which is not a normal position arrangement. Similarly, in the 5th column, sg2im predicts that ``pant'' is not vertically in line with the ``shirt''. Moreover, in the last column sg2im predicts that the scale of ``sky'' is too small.
It leads to the chaotic color fill in the generated image. Similarly, in the 7th column sg2im predicts the scale of ``snow'' is much bigger than ``mountain'', which conflicts with the triplet ``snow on mountain''. These displacements occur when the training process is not enhanced with relative distance and scales. 

Fig. \ref{fig:cmp_gt} shows the generated images conditioned on ground truth layouts. We compared our model grounded on the same position layouts compared with sg2im and LostGAN. It can be found that our method is more likely to generate realistic images from rich layouts and with natural objects. 

\noindent\textbf{User Study}\quad We also conduct a user study to measure human preference between images generated by out method and sg2im in Table~\ref{tab:user_study}. We choose the 128$\times$128 resolution models for both cases. 
Our user study involves 40 students having a background in computer science. We generate 500 test cases from the VG test set for user study. 
A majority of users preferred the generated layouts and images from our method in 65\% of image pairs.


%% file: content/5conclusion.tex
\section{Conclusion}
The relationship between objects significantly rectifies the localization of objects and even their appearances. Prior literature mainly focuses on fitting the single object appearance. Semantic interactions among objects were overlooked, which may result in inconsistent and chaotic results. In this paper, we proposed a new framework to generate complex-scene images 
by exploring the importance of relationships among multiple objects in complex-scene image generation. Quantitative results, qualitative results, and user studies show our method's ability for reasonable layout generation and object interactions alignment.

%% file: content/impact.tex
\section*{Ethical Impact}
Images are tiny visual samples of the grand physical world. The elementary particles that comprise our world first evolve and cluster into objects. Then appearance of objects are gradually shaped by the interactions between their counterparts. To model the natural clustering and interaction, we construct a generative model that strictly and structurally mimic the graph-like natural arrangement of our world. Our model builds a projection from symbolic graph space to the pixel space and we demonstrate how a small alternation in the object relationship can greatly affect the appearance of surrounding objects. In the future, we should research more on the neural design that can easily fit the data structure of pixels and encode more in the model visually commonsensical (relational) patterns. Overall, this paper has a positive impact on both industry and academia and enhance people's understanding of visual thinking. 